%% file: main.tex
\documentclass[10pt, a4paper]{article}

\usepackage[final]{main} 

\usepackage{amsmath}
\usepackage{graphicx}
\usepackage{float} 
\usepackage{subfigure}
\usepackage{multirow}
\usepackage{booktabs}
\usepackage{tcolorbox}
\usepackage{color}
\usepackage{pifont}
\usepackage{makecell}
\usepackage{verbatim}
\usepackage{colortbl}
\usepackage{microtype}
\usepackage{arydshln}

\usepackage{xcolor}
\usepackage{hyperref}
 \definecolor{darkblue}{rgb}{0, 0, 0.5}
  \hypersetup{colorlinks=true, citecolor=darkblue, linkcolor=darkblue, urlcolor=darkblue}
\usepackage{xstring}

\newcommand{\mask}{\textsc{[Mask]}}
\newcommand{\rel}[1]{\verb~#1~}
\newcommand{\ent}[1]{\textsc{#1}}

\def\sblue#1{\textcolor[rgb]{0,0,1}{\small~(#1)}}
\def\sred#1{\textcolor[rgb]{1,0,0}{\small~(#1)}}
\def\huggingface{\raisebox{-0.55ex}{\includegraphics[width=1.3em]{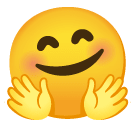}}}

\title{Take Care of Your Prompt Bias! Investigating and Mitigating Prompt Bias in Factual Knowledge Extraction}

\name{
Ziyang Xu${}^{1}$, 
Keqin Peng${}^3$, 
Liang Ding${}^4$\sthanks{~~Corresponding Authors: Liang Ding and Xiliang Lu.},
Dacheng Tao${}^5$, 
Xiliang Lu${}^{1,2,6*}$
}
\address{${}^1$Institute of Artificial Intelligence, School of Computer Science, 
Wuhan University, China \\
${}^2$School of Mathematics and Statistics, Wuhan University, China \\
${}^6$Hubei Key Laboratory of Computational Science, Wuhan University, China\\
${}^3$Beihang University,~${}^4$The University of Sydney,~${}^5$Nanyang Technological University\\
\texttt{\{xuziyang,xllv.math\}@whu.edu.cn}\\
\texttt{\{keqin.peng\}@buaa.edu.cn}\\
\texttt{\{liangding.liam,dacheng.tao\}@gmail.com}
}

\abstract{
Recent research shows that pre-trained language models (PLMs) suffer from ``prompt bias'' in factual knowledge extraction, i.e., prompts tend to introduce biases toward specific labels. Prompt bias presents a significant challenge in assessing the factual knowledge within PLMs. Therefore, this paper aims to improve the reliability of existing benchmarks by thoroughly investigating and mitigating prompt bias.  We show that: 1) all prompts in the experiments exhibit non-negligible bias, with gradient-based prompts like AutoPrompt and OptiPrompt displaying significantly higher levels of bias; 2) prompt bias can amplify benchmark accuracy unreasonably by overfitting the test datasets, especially on imbalanced datasets like LAMA.
Based on these findings, we propose a representation-based approach to mitigate the prompt bias during inference time. Specifically, 
we first estimate the biased representation using prompt-only querying, and then remove it from the model's internal representations to generate the debiased representations, which are used to produce the final debiased outputs.
Experiments across various prompts, PLMs, and benchmarks show that our approach can not only correct the overfitted performance caused by prompt bias, but also significantly improve the prompt retrieval capability (up to 10\% absolute performance gain). These results indicate that our approach effectively alleviates prompt bias in knowledge evaluation, thereby enhancing the reliability of benchmark assessments. Hopefully, our plug-and-play approach can be a golden standard to strengthen PLMs toward reliable knowledge bases. Code and data are released in \hyperlink{https://github.com/FelliYang/PromptBias}{https://github.com/FelliYang/PromptBias}.
\\ \newline }

\begin{document}
\maketitleabstract
%

\input{sec/1introduction.tex}

\input{sec/2prompt_bias.tex}

\input{sec/3method.tex}

\input{sec/4setup}

\input{sec/5experiments}

\input{sec/6analysis}

\input{sec/7relatedwork}
\input{sec/8conclusions}

\section*{Limitations}
Our debiasing method performs a preliminary exploration of how to use representation to mitigate bias, which we believe can be further explored and improved. Moreover, the prompt bias estimation strategy (by masking the input information) may be suboptimal. Additional strategies and analyses need to be proposed to better estimate the prompt bias. We leave these works for the future.

\section*{Ethics Statement}
We take ethical considerations very seriously and strictly adhere to the Ethics Policy. This paper focuses on mitigating prompt bias in factual knowledge extraction. The datasets used in this paper are publicly available and have been widely adopted by researchers. We ensure that the findings and conclusions of this paper are reported accurately and objectively.

\section*{Acknowledgements }
We are grateful to the anonymous reviewers and the area chair for their insightful comments and suggestions. This work is supported by the National Key Research and Development Program of China (No. SQ2023YFA1000103) and the National Nature Science Foundation of China (No.12371424)

\section*{Bibliographical References}

\bibliographystyle{main}
\bibliography{main}

\bibliographystylelanguageresource{main}
\bibliographylanguageresource{languageresource}

\appendix

\input{sec/9appendix.tex}

\end{document}

%% file: sec/1introduction.tex
\section{Introduction}

 Extracting factual knowledge from PLMs brings new vitality to the knowledge base construction community that typically requires high amounts of manual work from domain experts.
 Researchers have been fascinated by probing factual knowledge in PLMs~\cite{petroni-etal-2019-language,jiang-etal-2020-how,kassner-schutze-2020-negated,zhong-etal-2021-factual,cao-etal-2021-knowledgeable}. One commonly employed approach involves using prompt-based querying to extract knowledge from PLMs, i.e., prompting PLMs to fill masked object slots.

 \begin{figure}[tpb]
\centering
\includegraphics[width=0.44\textwidth]{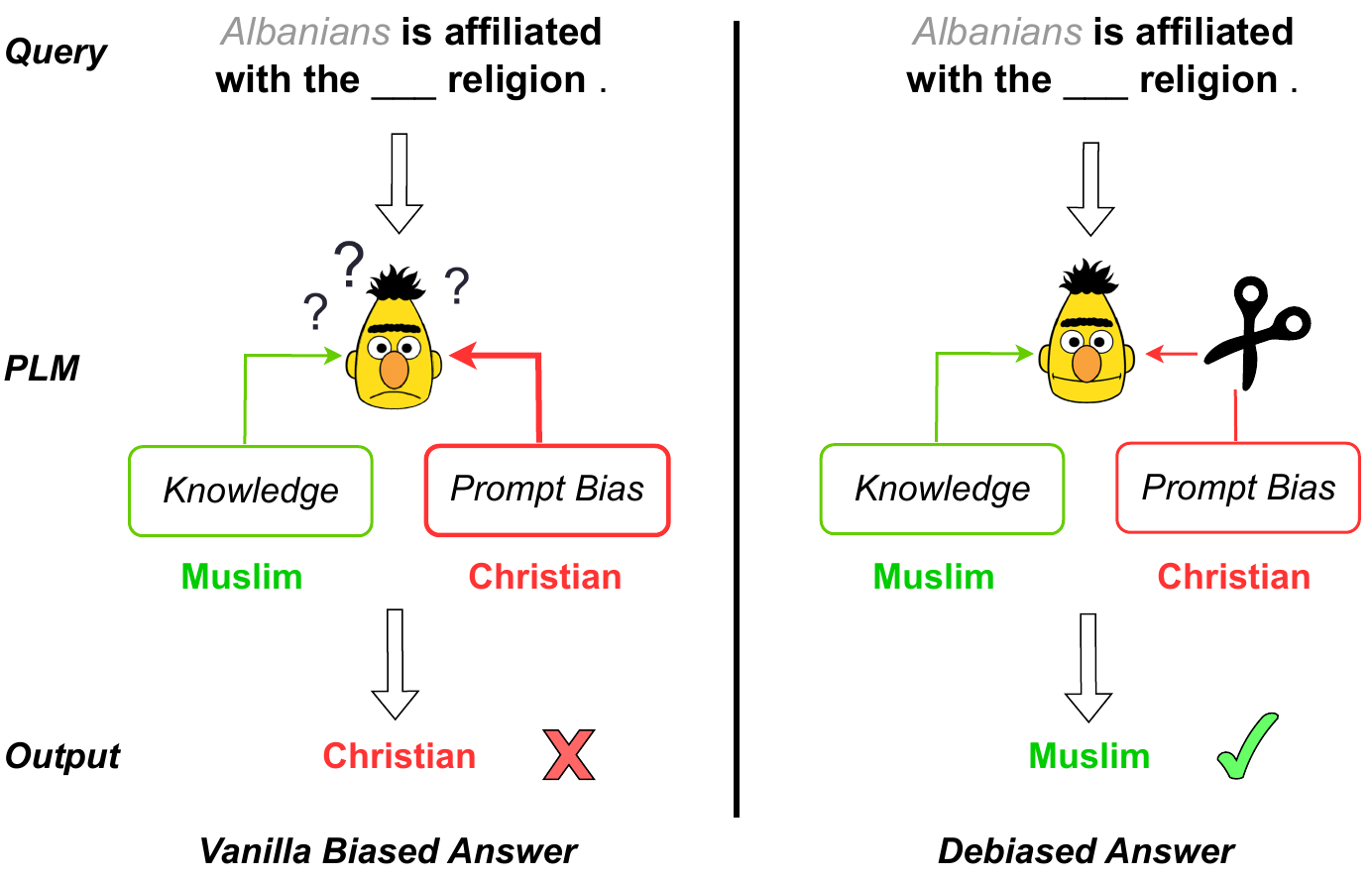}
\caption{Language models surfer from unintended \textbf{prompt bias} in factual knowledge extraction. When querying BERT the religion of \textit{Albanians}, the model is affected by the prompt bias and makes an incorrect prediction \textit{Christian}. With our debiasing approach, the model rectifies its prediction to the correct answer \textit{Muslim}.}
\label{fig:motivation}
\end{figure}

 However, recent research found that the outputs of prompt-based querying are dominated by prompt bias rather than PLMs' internal knowledge~\cite{cao-etal-2021-knowledgeable}, which strongly questions the reliability of current factual benchmarks. The interference of prompt bias makes it challenging to evaluate the amount of factual knowledge inside PLMs, significantly hindering language models from serving as reliable knowledge bases. 

 This paper aims to improve the reliability of factual knowledge benchmarks by thoroughly conducting a comprehensive analysis of prompt bias impact and mitigating it in the knowledge retrieval process. In this paper, we propose a general method to quantify the prompt bias across various PLMs and prompt types, and thoroughly assess its impact on widely used benchmarks such as LAMA. Experiments show that all current knowledge-extraction prompts have significant prompt bias, with previously reported high-performing prompts such as AutoPrompt and Optiprompt often exhibiting more pronounced bias. Moreover, we find that prompt bias severely compromises benchmark reliability by overfitting the test datasets and amplifying performance unreasonably. For example, the prompt bias of OptiPrompt helps inflate absolute accuracy by over \textbf{16.47}\% on the LAMA benchmark when probing the BERT-base model. Additionally, prompt bias can mislead language models and discourage models from making correct predictions, as illustrated in Figure~\ref{fig:motivation}. This significantly hinders the knowledge retrieval capabilities of prompts.

Based on these findings, we propose a representation-based approach to mitigate the prompt bias. Specifically,  we first construct a prompt-only query by replacing the subject slot with a meaningless subject such as "[MASK]". By leveraging the prompt-only query, we can estimate the PLM's bias toward the prompt, from which we generate a ``biased representation''. Subsequently, we utilize the biased representation to mitigate prompt bias by vector operations in the representation layer. \looseness=-1 

We conduct experiments across different prompts, PLMs and benchmarks, and find that our approach effectively rectifies the issue of inflated performance caused by prompt bias overfitting, thereby enhancing the reliability of factual knowledge benchmarks. Furthermore, our debiasing approach consistently and significantly improves the retrieval capability of prompts.


\begin{figure*}
\centering  
\includegraphics[width=0.94\textwidth]{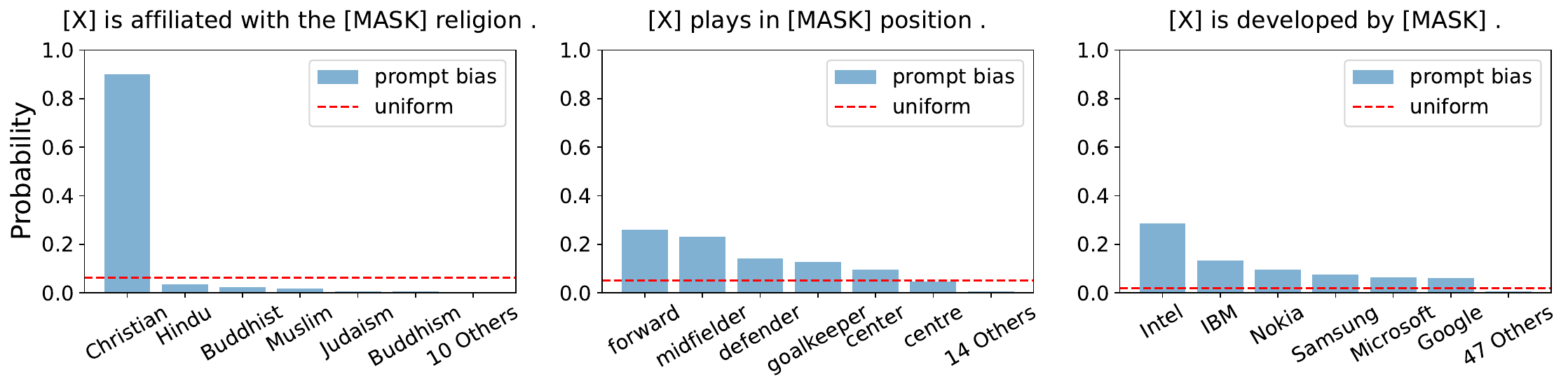}
\caption{Examples of prompt bias from LAMA manual prompts.  ``Prompt bias'' shows the BERT-base model probability distributions probed using prompt-only querying, while  ``uniform'' shows an ideal unbiased distribution for reference. \textbf{Prompts are biased towards certain labels.}}    
\label{fig:prompt_bias_examples}
\end{figure*}

Our contributions are as follows:

    
\begin{itemize}
    \item \textbf{We propose an approach to quantify the prompt bias in knowledge extraction and assess its impact on the reliability of benchmarks.} We show that all kinds of prompts in the experiments exhibit non-negligible prompt bias, with gradient-based prompts playing significantly higher levels of bias. Furthermore, we demonstrate two negative impacts of prompt bias: 1) unreasonably amplifying benchmark performance through overfitting; 2) impairing the prompt retrieval capability by misleading PLMs.
    \item \textbf{We propose a representation-based debiasing approach to effectively tackle the challenge of prompt bias. }Experiments show that our approach effectively rectifies the inflated performance caused by overfitting and improves prompt retrieval capability (up to 10\% absolute), presenting a reliable and better performance.
    \item \textbf{After mitigating prompt bias, we observe that the knowledge retrieval abilities of manually designed prompts are comparable to or better than those of state-of-the-art prompts.} Our debiasing approach has shed light on the actual retrieval capabilities of these more complex prompts, where we do not observe significant performance improvements. OptiPrompt stands out as an exception, although its debiased retrieval performance still falls short of expectations. 
\end{itemize}

This paper is an early step in exploring reliable factual knowledge extraction from PLMs, which employs a simple but effective debiasing approach. 
We recommend improving the faithfulness of existing prompt-based factual knowledge extraction approaches using debiasing methods.

%% file: sec/2prompt_bias.tex


\section{Investigating Prompt Bias}
\label{sec:prompt_bias}

\subsection{Uncovering Prompt Bias}
\label{sec:observe_bias}
Prompts are crucial in converting the downstream task format into a natural language format that PLMs can understand. In this process, an ideal prompt, which typically specifies the label space, should not be inherently biased toward any particular label.  However, in practice, prompts can introduce bias towards specific labels, as observed in tasks like factual knowledge extraction~\citet{cao-etal-2021-knowledgeable}, which we term as \textbf{prompt bias}. 

To demonstrate the bias induced when using prompts to probe factual knowledge, we follow previous works~\cite{zhao-etal-2021-calibrate, cao-etal-2021-knowledgeable} to use \textit{prompt-only querying} to probe PLMs. Specifically, prompt-only querying constructs a special input for each prompt by replacing the \rel{subject} slot [X] in a prompt~(e.g., ``[X] used to communicate in \mask.'') with a meaningless token such as \mask, N/A or "". By employing prompt-only querying for probing PLMs, we can observe the inherent bias of the prompt.

Following~\citet{cao-etal-2021-knowledgeable}, we use \mask~as the meaningless token in this paper. 
Ideally, in the absence of valid subject information, prompt-only querying should exhibit a uniform distribution within the label space. However, the prompts employed in probing factual knowledge show a severe bias towards specific labels. Figure~\ref{fig:prompt_bias_examples} shows the output distributions of prompt-only querying for several prompts in the LAMA benchmark.  For example, when probing with ``[X] is affiliated with the \mask~religion.'', the BERT-base model is severely biased towards \rel{Christian}, showing a probability as high as 90\%. The prompt bias also exists in other fields such as sports and industry, as shown in Figure~\ref{fig:prompt_bias_examples}. 

\subsection{Quantifying Prompt Bias}
To comprehensively investigate the prompt bias across different types of prompts and language models, we quantify the prompt bias using the Jensen–Shannon(J-S) divergence, which is derived from the Kullback–Leibler divergence and addresses its asymmetry and infinite value range.

Specifically, for a language model $M$ and a prompt $T$, we define the output probabilities of prompt-only querying as the \textit{prompt bias distribution}.
Then we quantify the prompt bias using the J-S divergence between the prompt bias distribution and the uniform distribution, formulated by:
\begin{align}
bias = JS(P_M(y|T), U) ,
\end{align}
where $P_M(y|T)$ and $U$ refer to prompt bias distribution and uniform distribution respectively. 

This measurement provides an intuitive quantification of the prompt bias, where a larger J-S divergence indicates a greater degree of prompt bias.

\begin{figure}[htpb]
\centering
\includegraphics[width=0.41\textwidth]{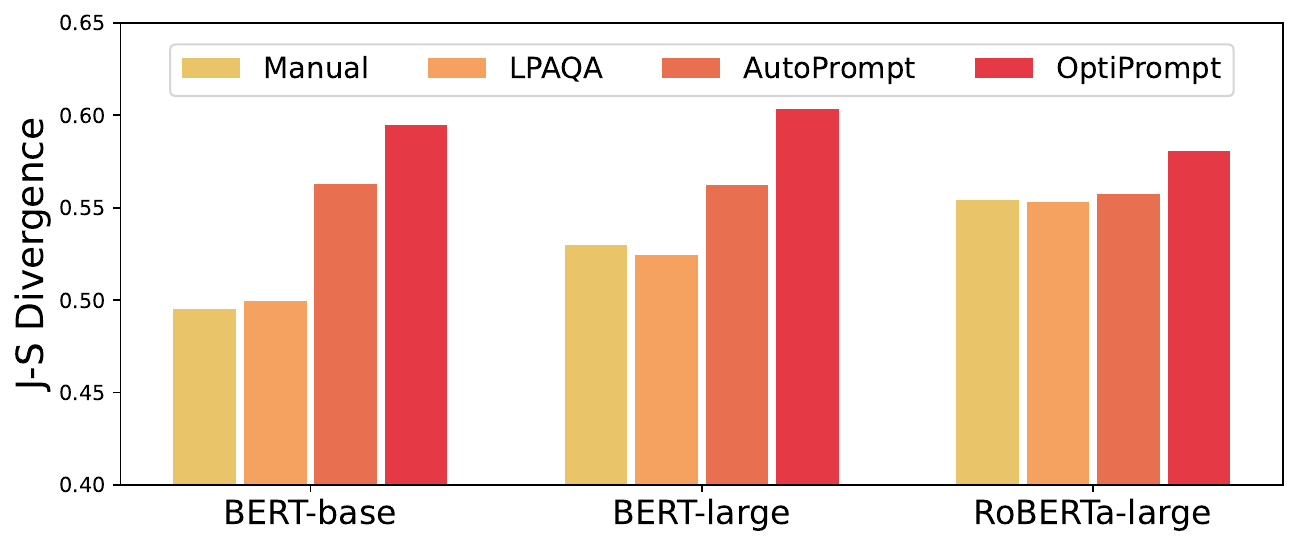}
    \caption{Quantified prompt bias for various prompts and PLMs using J-S divergence, averaged on 41 relations in the LAMA benchmark.
    }
    \label{fig:quantify_bias}
\end{figure}

As shown in Figure~\ref{fig:quantify_bias}, we quantify the prompt bias of four different prompts across three PLMs. The bias is averaged over 41 relations in the LAMA benchmark. For additional details about the prompts and benchmark employed in our study, please refer to Section~\ref{sec:exp_setup}.

Notably, we observe significant prompt bias across all prompt types and PLMs in our experiments. Specifically, concerning prompt types, manual prompts and paraphrase-based prompts (LPAQA) exhibit a comparable degree of bias, whereas gradient-based prompts (AutoPrompt and OptiPrompt) show a more pronounced bias,  up to 0.6 for OptiPrompt.
Additionally, although OptiPrompt and AutoPrompt use the same training dataset, OptiPrompt exhibits a higher level of bias compared to AutoPrompt. This may stem from OptiPrompt's more comprehensive optimization within the continuous embedding space.

Regarding different PLMs, BERT-large exhibits a greater bias than BERT-base, especially in manual and paraphrase-based prompts. 
Furthermore, the bias exhibited by BERT models varies significantly across different prompts, whereas the bias of RoBERTa remains relatively stable, consistently ranging between 0.55 and 0.58.

\subsection{Prompt Bias Impact on Benchmarks}
Although we have previously observed and quantified prompt bias, it remains unclear how prompt bias affects benchmark evaluation. This section demonstrates two negative impacts of prompt bias.

\paragraph{Prompt Bias Can Overfit Datasets and Impairs Benchmark Reliability.}
 We first assess the overfitting degree of prompt bias on benchmark datasets. 
 To address this, we explore two strategies for leveraging prompt bias to answer factual knowledge queries within the benchmark. The first strategy keep predicting the label with the highest probability within the prompt bias distribution, e.g., keep predicting \rel{Christian} for all queries in the LAMA benchmark. The second 
 strategy involves sampling predictions from the prompt bias distribution. We employ the first strategy in this experiment due to the fact that it shows a larger overfitting performance in practice.

Table~\ref{tab:prompt_accuracy} presents the results, with prompts arranged based on their vanilla performance (see Table \ref{tab:prompt_based} for additional information). Surprisingly, across various prompts and PLMs, prompt bias achieves non-trivial performance on LAMA and LAMA-UHN,  which are widely used benchmarks with imbalanced data distributions. Additionally, gradient-based prompts exhibit a more pronounced overfitting compared to manual and paraphrased prompts.
In contrast, prompt bias exhibits limited influence on WIKI-UNI due to the uniform data distribution characteristic of this benchmark.
\begin{table}[htpb]
    \centering
    \resizebox{\linewidth}{!}{
    \begin{tabular}{lccc}
    \toprule
        \textbf{Prompts} & \textbf{LAMA} & \textbf{LAMA-UHN} & \textbf{WIKI-UNI} \\
    \midrule
         Manual & 5.23 & 4.87 & 1.05 \\
        LPAQA & 6.36 & 5.89 & 1.7 \\
        AutoPrompt & 13.52 & 13.16 & 1.71 \\
        OptiPrompt & 16.47 & 17.55 & 1.72 \\
    \bottomrule
    \end{tabular}
    }
    \caption{The overfit accuracy of four types of prompts on three test datasets, probed with BERT-base, averaged across 41 relations for each dataset.  The accuracy is assessed by always predicting the label biased most by the prompt. \textbf{The prompt bias severely overfits imbalanced datasets like LAMA and LAMA-UHN.}}
    \label{tab:prompt_accuracy}
\end{table}

\begin{table}[htpb]
    \centering
    \resizebox{\linewidth}{!}{
    \begin{tabular}{lccc}
    \toprule
        \textbf{Prompts} & \textbf{LAMA} & \textbf{LAMA-UHN} & \textbf{WIKI-UNI} \\
    \midrule
         Manual & 37.1 & 27.3 & 20.0 \\
        LPAQA & 38.1 & 29.0 & 19.7 \\
        AutoPrompt & 43.9 & 33.4 & 20.6 \\
       OptiPrompt & 49.4 & 39.5 & 23.1 \\
    \bottomrule
    \end{tabular}
    }
    \caption{Prompts' vanilla performance on three test datasets, probed with BERT-base.}
    \label{tab:prompt_based}
\end{table}

\paragraph{Prompt Bias Can Mislead Language Models and Impair the Prompt Retrieval Capability.} In addition to overfitting benchmarks, we observe that prompt bias can potentially mislead PLMs in factual knowledge probing. Table~\ref{tab:rectified_examples} illustrates this phenomenon using examples from the LAMA benchmark. For instance, when querying the BERT-base model with ``Albanians is affiliated with the \mask~religion .'', it will be affected by the strong prompt bias shown in Figure~\ref{fig:prompt_bias_examples}
and predict the incorrect label \rel{Christian}. However, the model can predict the correct answer \rel{Muslim}~after mitigating prompt bias using the approach introduced in~Section\ref{sec:method}. This suggests that prompt bias can mislead PLMs and prevent them from fully using their knowledge to answer factual queries.

As a result, besides lowering the reliability of benchmarks, prompt bias also impairs the knowledge retrieval capability of prompts.

\newcommand{\WrongPred}[1]{\textcolor{red}{#1}}
\newcommand{\RightPred}[1]{\textcolor{blue}{#1}}

\begin{table}[tbp]
    \centering
    \renewcommand{\arraystretch}{2} 
    {\fontsize{1.5\baselineskip}{1.5\baselineskip}\selectfont
    \resizebox{\linewidth}{!}{
    
    \begin{tabular}{lcc}
        \toprule
        \textbf{Example} & \textbf{Vanilla} & \textbf{Debiased} \\
        \midrule
        Albanians is affiliated with the \mask~religion .& \WrongPred{Christian}& \RightPred{Muslim}\\
        Afghanistan is affiliated with the \mask~religion .& \WrongPred{Christian}& \RightPred{Islam}\\
        \hdashline
        
        	Vladislav Tretiak plays in \mask~position .& \WrongPred{midfielder}& \RightPred{goaltender}\\
          Tuukka Rask plays in \mask~position .& \WrongPred{forward}& \RightPred{goaltender}\\
        \hdashline
 iChat is developed by \mask~.& \WrongPred{Intel}& \RightPred{Apple}\\
        Digital Negative is developed by \mask~.& \WrongPred{Intel}& \RightPred{Adobe}\\
        
        \bottomrule
    \end{tabular}
    }
    }
    \caption{Examples in the LAMA benchmark where BERT-base is misled by prompt bias and makes incorrect predictions. After debiasing, the model gives correct answers. Vanilla and rectified predictions are shown in \WrongPred{red} and \RightPred{blue} respectively.}
    \label{tab:rectified_examples}
\end{table}





\begin{figure*}[htbp] 
\centering 
\includegraphics[width=0.94\textwidth]{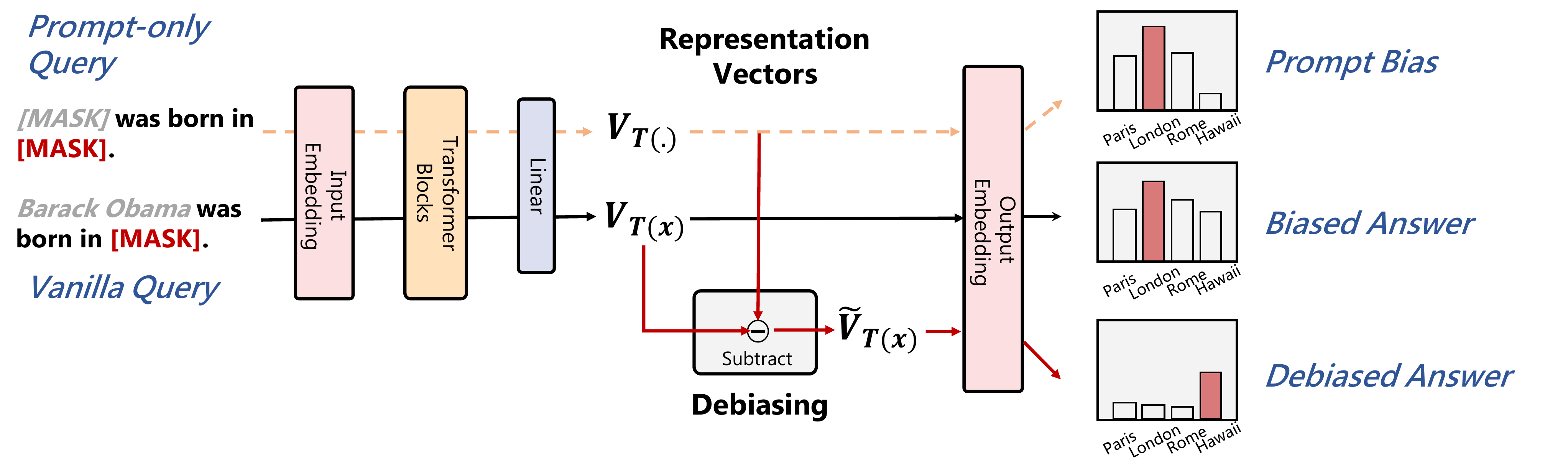} 
\caption{
The pipeline of our method. The red line represents the process of debiasing, which uses the subtraction of the representations between prompt-only query $V_{T(\cdot)}$ and vanilla query $V_{T(x)}$.
}
\label{fig:model} 
\end{figure*}

%% file: sec/3method.tex
\section{Mitigating Prompt Bias}
\label{sec:method}





Thus far, we have shown that PLMs suffer from severe prompt bias in factual knowledge extraction, which negatively impairs benchmark reliability as well as prompt retrieval capability.
Here, we propose a representation-based approach to mitigate prompt bias. The internal representation vectors of PLMs encompass their preference for answer labels, which incorporate both internal knowledge and prompt bias. Therefore, the prompt bias can be addressed by removing the biased component from the representation vectors. The key idea is to identify the biased component in representations through prompt-only querying. Figure~\ref{fig:model} shows the pipeline of our approach.


In detail, given a prompt $T$ (e.g., ``[X] was born in \mask.'') and a fact knowledge subject $x$, we first construct the vanilla query $T(x)$ and the prompt-only query $T(\cdot)$ by replacing the "[X]" slot in $T$ with $x$ and a meaningless token like \mask~, respectively. Then we push $T(x)$ and $T(\cdot)$ into PLMs to get their representation vectors on the masked position, denoted as $V_{T(x)}$ and $V_{T(\cdot)}$. Specifically, the representation vectors refer to the outputs of the model's final layer\footnote{The final layer refers to the layer preceding the output embedding layer. Typically, the final layer corresponds to the last transformer block, but it may also denote the linear layer following the transformer blocks in some cases such as in BERT models.}, which is crucial for the debiasing algorithm to work\footnote{Outputs of the final layer don't suffer from non-linear transformations until being decoded into tokens. Therefore, they serve as a solid foundation for the subsequent linear operations involved in the debiasing process. }.
Next, we take the subtraction of two representation vectors as the debiased vector, namely~$\tilde{V}_{T(x)}$. This process is formulated by:
\begin{align}
    \tilde{V}_{T(x)} &= V_{T(x)} - V_{T(\cdot)}.
    \label{eq:1}
\end{align}

We then use $\tilde{V}_{T(x)}$ to replace the original representation vector $V_{T(x)}$ in the decoding stage and get the debiased logits:
\begin{align}
    \tilde{L}_{T(x)} &=  E^o(\tilde{V}_{T(x)}),
\end{align}
where $E^o$ represents the output embedding layer of the PLM. 
To obtain the final prediction, we select the label with the highest logit value in the label space $\mathcal{C}$ (see Section~\ref{subsec:querytype} for more details) : 
\begin{align}
    \text{token} = \arg\max_{v \in \mathcal{C}} \tilde{L}_{T(x)}(v).
\end{align}

Unlike previous approaches that adjust output probabilities using an affine transformation~\cite{guo-etal-2017-on,zhao-etal-2021-calibrate}, we address prompt bias in the representation layer. One advantage of our approach lies in its capability to generate debiased representation vectors, which can be utilized to mitigate bias in broader scenarios, such as sentence embeddings~\citep{jiangPromptBERTImprovingBERT2022}.

%% file: sec/4setup.tex
\section{Experimental Setup}
\label{sec:exp_setup}
\subsection{Model Details}
Following previous work~\cite{zhong-etal-2021-factual}, We conducted our main experiments on BERT-base, BERT-large~\cite{devlin-etal-2019-BERT} and RoBERTa-large~\citep{liu-etal-2020-RoBERTa} models, which are widely used in factual knowledge extraction task. Additionally, we perform experiments on Llama2~\cite{touvron2023llama} 7B to further investigate the generalizability of the debiasing strategy.

\subsection{Prompt Settings}
We conducted experiments using four different types of prompts:

\noindent\textbf{LAMA Manual} refers to a series of manual prompts constructed by \citet{petroni-etal-2019-language} to probe factual knowledge within PLMs. 
To extract factual knowledge from PLMs, \citet{petroni-etal-2019-language} manually constructed a specific prompt for each relation in the LAMA benchmark. For example, the prompt ``[X] was born in \mask.'' is designed for the relation \rel{place-of-birth}. 
 
\noindent\textbf{LPAQA} refers to the mining-based and paraphrasing-based prompts constructed by~\citet{jiang-etal-2020-how}, which demonstrate better performance than LAMA Manual in the LAMA benchmark.

\noindent\textbf{AutoPrompt} refers to the gradient-guided prompts that are optimized on discrete token space by~\citet{Shin-etal-2020-Auto}. Additional training datasets are required for optimizing the prompts. 

\noindent\textbf{OptiPrompt} refers to the continuous prompts proposed by~\citep{zhong-etal-2021-factual}, which optimizes in the continuous embedding space using the same training datasets as AutoPrompt. 

In our experiments, we directly utilize the prompts published by these works. An exception is OptiPrompt; due to the lack of officially available prompts, we train OptiPrompt for BERT and RoBERTa models according to the settings outlined in the paper~\citet{zhong-etal-2021-factual}. Further optimization details can be found in Appendix~\ref{apx:optimization}.

\subsection{Benchmarks}
We involve three benchmarks: two imbalanced benchmarks (LAMA and LAMA-UHN), and one balanced benchmark (WIKI-UNI). The test datasets of balanced benchmarks have a uniform label distribution, unlike those of imbalanced benchmarks. 

\noindent\textbf{LAMA} is a widely-used benchmark, originally constructed by \citet{petroni-etal-2019-language}, designed to evaluate various knowledge contained in PLMs. In LAMA, a fact is defined as a triple (\ent{subject}, \rel{relation}, \ent{object}) such as (\ent{Dante}, \rel{born-in}, \ent{Florence}). Following previous work~\citet{zhong-etal-2021-factual}, we focus on factual knowledge extraction and use the TREx~\cite{Elsahar-etal-2018-rex} subset of LAMA in the experiments. It contains up to 1000 fact triples for each of the 41 Wikidata relation types. Notably, LAMA is an imbalanced dataset, particularly regarding certain relations such as P136 \rel{genre}, where the label ``\ent{Jazz}'' constitutes over 70\% of the dataset.

\noindent\textbf{LAMA-UHN} is a more challenging variant of LAMA constructed by \citet{poerner-etal-2020-e}, which is also imbalanced. In comparison to LAMA, LAMA-UHN filters out fact triples that are easy to guess, such as cases where the object is a substring of the subject.

\noindent\textbf{WIKI-UNI} is a balanced dataset constructed from Wikidata~\cite{cao-etal-2021-knowledgeable}.  It has been meticulously curated to ensure a uniform answer distribution. WIKI-UNI encompasses the same 41 relations as LAMA and is of comparable size.


\subsection{Querying Paradigms} 
\label{subsec:querytype}
In factual knowledge extraction, there are mainly two types of querying paradigms:
\begin{itemize}
\item \textbf{Untyped querying}~\citep{petroni-etal-2019-language, jiang-etal-2020-how, zhong-etal-2021-factual} involves querying PLMs for object answers in the whole vocabulary or inter-vocab of different PLMs.
\item \textbf{Typed querying}~\citep{kassner-etal-2021-multilingual, xiong-etal-2020-Pretrained} involves querying PLMs for object answers in a candidate set $\mathcal{C}$ that consists of expected type tokens. For example, the candidate set $\mathcal{C}$ for templates such as "[X] was born in \mask." includes all cities present in the PLM's vocabulary.
\end{itemize}

In this paper, we focus on prompt bias in typed querying. We take the candidate set of previous work~\citep{kassner-etal-2021-multilingual} as a basis and expand it by adding extra labels inside the test datasets. Additionally, to maintain consistency with previous work~\citep{cao-etal-2021-knowledgeable,zhong-etal-2021-factual}, we only allow candidate labels consisting of a single token for BERT and RoBERTa models. Our code is implemented using \huggingface~Transformers~\citep{wolf-etal-2019-tranformers} and OpenPrompt~\citep{ding-etal-2022-openprompt}.

\def\biasedColor{\rowcolor{red!5}}
\def\fairColor{\rowcolor{green!5}}


\begin{table*}[htbp]
    \centering
    \resizebox{0.91\linewidth}{!}{
    \begin{tabular}{llcllcllcl}
        \toprule
        \multirow{2}*{\textbf{Datasets}} & \multirow{2}*{\textbf{Prompts}} & \multicolumn{2}{c}{\textbf{BERT-\textsc{Base}}} & & \multicolumn{2}{c}{\textbf{BERT-\textsc{Large}}} & & \multicolumn{2}{c}{\textbf{Roberta-\textsc{Large}}} \\
        ~ & ~ & \makecell[c]{\textbf{Prec.}} & \makecell[c]{\textbf{Prec.$^d$}} & & \makecell[c]{\textbf{Prec.}} & \makecell[c]{\textbf{Prec.$^d$}} & & \makecell[c]{\textbf{Prec.}} & \makecell[c]{\textbf{Prec.$^d$}}\\
\hline
\fairColor & Manual & 20.0 & 24.2\sred{ +4.2 } & & 22.6 & 26.1\sred{ +3.5 } & & 20.1 & 24.0\sred{ +3.9 }\\
\fairColor & LPAQA & 19.7 & 24.3\sred{ +4.6 } & & 21.6 & 24.4\sred{ +2.8 } & & 20.1 & 23.6\sred{ +3.5 }\\
\fairColor & AutoPrompt & 20.6 & 24.7\sred{ +4.1 } & & 21.2 & 25.0\sred{ +3.8 } & & 18.8 & 23.4\sred{ +4.6 }\\
\fairColor \multirow{-4}*{ WIKI-UNI } & OptiPrompt & 23.1 & 25.7\sred{ +2.6 } & & 24.8 & 28.3\sred{ +3.5 } & & 22.2 & 26.4\sred{ +4.2 }\\
\hline
\biasedColor & Manual & 37.1 & 32.4\sblue{ -4.7 } & & 38.7 & 32.2\sblue{ -6.5 } & & 36.4 & 30.7\sblue{ -5.7 }\\
\biasedColor & LPAQA & 38.1 & 31.0\sblue{ -7.1 } & & 40.2 & 31.6\sblue{ -8.6 } & & 39.0 & 30.5\sblue{ -8.5 }\\
\biasedColor & AutoPrompt & 43.9 & 33.4\sblue{ -10.5 } & & 43.5 & 34.8\sblue{ -8.7 } & & 42.3 & 26.1\sblue{ -16.2 }\\
\biasedColor \multirow{-4}*{ LAMA } & OptiPrompt & 49.7 & 34.1\sblue{ -15.6 } & & 52.4 & 38.6\sblue{ -13.8 } & & 48.5 & 34.7\sblue{ -13.8 }\\
\hline
\biasedColor & Manual & 27.2 & 22.5\sblue{ -4.7 } & & 30.0 & 23.0\sblue{ -7.0 } & & 28.3 & 22.0\sblue{ -6.3 }\\
\biasedColor & LPAQA & 29.0 & 21.2\sblue{ -7.8 } & & 31.7 & 22.5\sblue{ -9.2 } & & 31.1 & 21.9\sblue{ -9.2 }\\
\biasedColor & AutoPrompt & 33.4 & 22.1\sblue{ -11.3 } & & 33.8 & 25.0\sblue{ -8.8 } & & 33.9 & 17.6\sblue{ -16.3 }\\
\biasedColor \multirow{-4}*{ LAMA-UHN } & OptiPrompt & 39.8 & 23.1\sblue{ -16.7 } & & 43.6 & 28.3\sblue{ -15.3 } & & 39.8 & 24.9\sblue{ -14.9 }\\
        
        \bottomrule
    \end{tabular}
    }
    \caption{Top 1 accuracy before and after debiasing across various PLMs, prompts, and benchmarks, averaged across 41 relations. Results on imbalanced datasets (LAMA, LAMA-UHN) and the balanced dataset (WIKI-UNI) are shown in red \colorbox[HTML]{fdcbcb}{\phantom{O}} and green \colorbox[HTML]{abf9ab}{\phantom{O}} backgrounds respectively. The superscript $d$ denotes the debiased performance.}
    
    \label{tab:main_result}
\end{table*}

%% file: sec/5experiments.tex
\section{Experiments and Results}
In this section, we evaluate the effectiveness of our debiasing approach across various prompts and PLMs. Results are reported in Table~\ref{tab:main_result}.
\paragraph{Our Approach Rectifies the Inflated Accuracy of Imbalanced Datasets.}

After applying the debiasing approach, all prompts suffer from varying degrees of performance degradation on imbalanced benchmarks, consistently across different PLMs. Prompts exhibiting significant overfitting in Table~\ref{tab:prompt_accuracy} experience greater performance degradation, by up to -16.7\% absolute.
We attribute the decline to the correction of overfitted performance caused by prompt bias. To validate this, we conduct a thorough analysis to identify the source of the degradation.

\begin{figure}[htpb]
\centering
\includegraphics[width=0.41\textwidth]{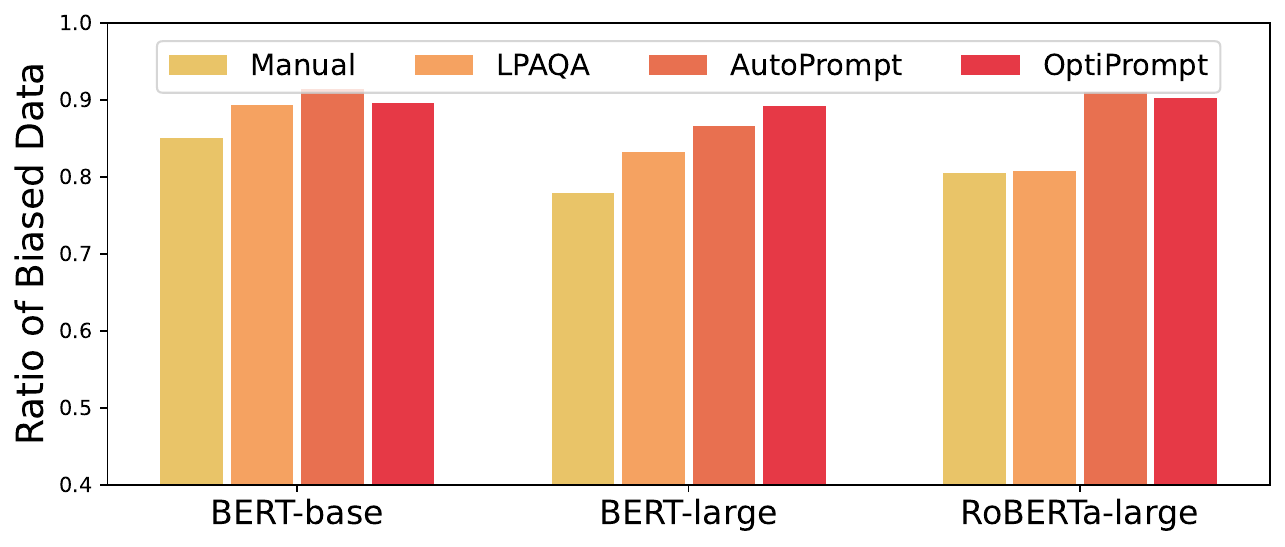}
    \caption{The ratio of biased data in the performance degraded by debiasing, across diverse PLMs and prompts, averaged on 41 relations. \textbf{Performance degradation mostly comes from biased data.}
    }
    \label{fig:bias_ratio}
\end{figure}

Specifically, we first select biased labels from the label space according to their probabilities in prompt-only querying. Biased labels refer to labels whose probabilities are higher than those of the uniform distribution ( e.g., \rel{Christian} in Figure~\ref{fig:prompt_bias_examples}). Then we identify test data containing biased labels as biased data, as they are prone to being overfitted by prompt bias.
Figure~\ref{fig:bias_ratio} plots the ratio of biased data among the incorrect predictions contributing to the performance degradation.

We find that incorrect predictions caused by the debiasing algorithm mostly come from biased data, accounting for up to approximately 90\% in both OptiPrompt and AutoPrompt. This suggests that the performance degradation is mainly attributed to performance correction for prompt bias overfit.
The analysis in Section~\ref{sec:analysis} further supports this.


\paragraph{Our Approach Improves the Accuracy of the Balanced Dataset.} 
Results of the balanced dataset WIKI-UNI listed in Table~\ref{tab:main_result} show that our approach consistently and significantly improves PLMs' accuracy on the WIKI-UNI benchmark, e.g., average +\textbf{3.9}, +\textbf{3.4} and +\textbf{4.1} upon BERT-base, BERT-large and RoBERTa-large, respectively. 
The improvement mainly comes from the rectification of incorrect predictions misled by prompt bias, as illustrated in Table~\ref{tab:rectified_examples}.


\begin{figure*}
\centering  
\includegraphics[width=0.94\textwidth]{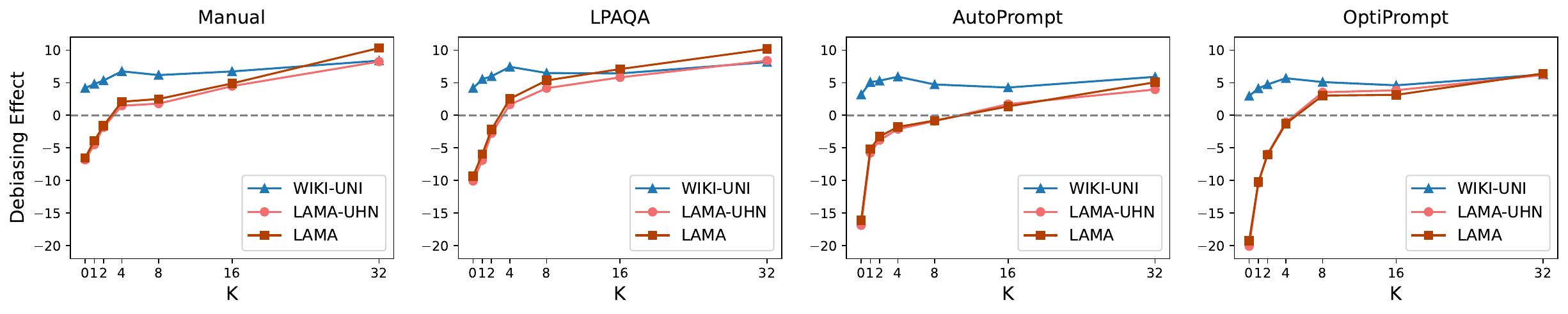}

\caption{The impact of debiasing on benchmark accuracy after filtering out prompt-overfitting data, probed on the BERT-base model using various prompts. K represents the number of biased labels filtered from raw benchmarks, see Section~\ref{sec:filter_out_data} for details. After filtering out overfitting data, \textbf{the debiasing impact on imbalanced datasets (LAMA, LAMA-UHN) turns from negative to positive, finally achieving comparable improvements with the balanced dataset (WIKI-UNI).}}    
\label{fig:filter_out_acc_variation}    
\end{figure*}

\paragraph{Limited Improvement in Knowledge Retrieval Abilities with More Complex Prompts. }
Previous studies have made efforts to seek prompts with better knowledge retrieval capability, such as LPAQA, AutoPrompt, and OptiPrompt. Despite demonstrating improved performance on benchmarks like LAMA,  these complex prompts typically exhibit larger overfitting, as evidenced in~Table\ref{tab:prompt_accuracy}. This raises doubts about their actual effectiveness in knowledge retrieval. 

Our debiasing algorithm sheds light on this issue. Upon mitigating prompt bias,  the performance of these prompts returns to the same level on imbalanced datasets.  Furthermore, their performance is close to each other on the balanced WIKI-UNI dataset, regardless of debiasing. Although OptiPrompt exhibits slightly better performance (e.g., average \textbf{+4.3} on BERT-large), it still falls short of expectations. These findings suggest that the purported “better” prompts' knowledge retrieval capability is not substantially enhanced when compared to manual prompts. 



%% file: sec/6analysis.tex
\section{Analysis and Discussion}
\label{sec:analysis}

\paragraph{Debiasing Shows Positive Impacts on Imbalanced Datasets After Filtering Out Prompt-Biased Data.}
\label{sec:filter_out_data}
In the main experiments, we have shown that the debiased performance drops dramatically on imbalanced datasets (LAMA, LAMA-UHN), which we attribute to our debiasing approach correcting the overfitting performance.
To further support our interpretation and explore more deeply the impact of prompt bias on imbalanced datasets, we design another experiment where we filter out data from imbalanced datasets that may be overfitted by the prompt bias, namely prompt-biased data. We are going to study how debiasing affects LAMA and LAMA-UHN performance without the interference of prompt-biased data.

Concretely, we first use prompt-only querying to probe the model and find the top-k labels biased by the prompt. Then we filter out data whose labels are in the top-k biased labels from the datasets. In our setting, $k$ could be 0, 1, 2, 4, 8, 16, 32. Table~\ref{tab:filter_out_datasize} shows the filtered dataset size for different $k$. 

Results on these filtered datasets are reported in Figure~\ref{fig:filter_out_acc_variation}.  
We only show the results on the BERT-base model in the main text.
Results on other models are reported in appendix \ref{apx:addition}.  
Notably, the performance of LAMA and LAMA-UHN exhibits a significant improvement after filtering out prompt-biased data.  With the increase of $k$, the debiasing effect on imbalanced datasets gradually shifts from negative to positive, eventually achieving a level of improvement comparable to that observed in the balanced dataset WIKI-UNI. These results indicate that our debiasing approach actually \textbf{improves} rather than degrades the retrieval ability of prompts on imbalanced datasets, by up to +10.8 on LAMA for the LPAQA prompt when k=32.

\begin{figure}[htpb]
\centering
\includegraphics[width=0.37\textwidth]{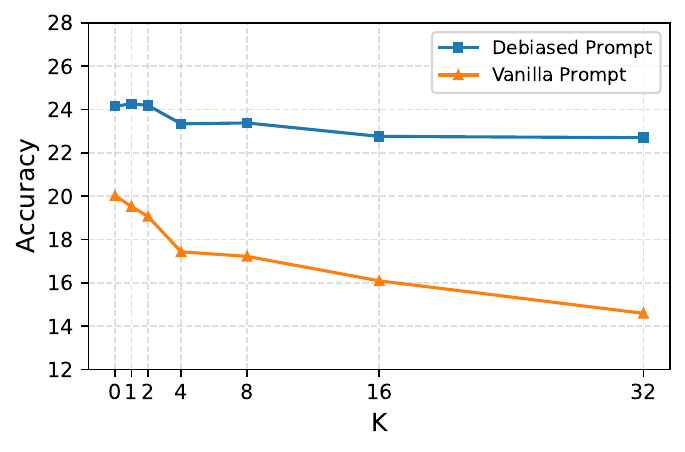}
    \caption{Manual prompts vanilla and debiased accuracy on filtered WIKI-UNI using different k, probed with BERT-base.}
        
    \label{fig:Manual_wiki_acc_var}
\end{figure}

\paragraph{Debiasing Benefit for Prompt Retrieval Capability is Underestimated. }
Another notable phenomenon is that the debiasing benefit shown on WIKI-UNI has further improved after removing prompt-biased data. For example, the debiasing improvements for Manual prompts increase from 4.1 (k=0) to 8.1 (k=32) on the BERT-base model, as shown in Figure~\ref{fig:filter_out_acc_variation}.  

To understand this phenomenon, we thoroughly analyze the performance variance on filtered WIKI-UNI. Results are reported in Figure~\ref{fig:Manual_wiki_acc_var}. We observe that vanilla performance drops significantly from 20.0~(k=0) to 14.5~(k=32) after removing prompt-biased data. This indicates that prompt bias also overfit WIKI-UNI, though not as high as LAMA and LAMA-UHN.
Interestingly, while the vanilla performance drops, the debiased performance is relatively stable. This suggests the benefit of our debiasing method may be underestimated, with the interference of prompt-biased data.

In summary, prompt bias can overfit a few data in the balanced dataset.  Using WIKI-UNI directly to evaluate the debiasing benefit on retrieval capabilities may lead to underestimation.

\paragraph{Generalizability of the Debiasing Algorithm on Large Large Models.}
We further evaluate the effectiveness of the debiasing strategy on Lama2 7B for Manual and LPAQA prompts. To adapt to auto-regressive language models, we make some adjustments to the debiasing algorithm; please refer to Appendix~\ref{apx:llama2} for more details. Table~\ref{tab:lama2} shows the results. According to the results, our debiasing strategy consistently enhances performance and reliability on Llama2, specifically by improving performance on WIKI-UNI (by up to +4.5) and rectifying overfitted performance on other imbalanced datasets. 
These results indicate that large language models are also at risk of suffering from prompt bias. Remarkably, the debiased performance of Llama2 7B significantly surpasses that of BERT and RoBERTa, which might be due to the fact that Llama2 contains much more factual knowledge.

Overall, our method demonstrates strong generalization capabilities and offers novel insights into addressing bias issues in large language models.

\begin{table}[htbp]
    \centering
    \begin{tabular}{lccc}
    \toprule
    \textbf{Datasets} & \textbf{Prompts} & \textbf{Prec.} & \textbf{Prec.$^d$}  \\
    \hline
     \fairColor & Manual & 27.5 & 31.9\sred{+4.4}\\
   \fairColor \multirow{-2}*{WIKI-UNI } & LPAQA & 26.4 & 30.9\sred{+4.5}\\
    \hline
     \biasedColor & Manual & 51.9 & 48.4\sblue{-3.5}\\
    \biasedColor \multirow{-2}*{LAMA} & LPAQA & 49.2 & 48.0\sblue{-1.2}\\
    \hline
    \biasedColor & Manual & 46.7 & 43.5\sblue{-3.2} \\
   \biasedColor \multirow{-2}*{LAMA-UHN} & LPAQA & 45.8 & 43.8\sblue{-2.0} \\
    \bottomrule
    \end{tabular}
    \caption{Top 1 accuracy before and after debiasing of Llama2 7B across several prompts and datasets, averaged on 41 relations. The symbols and color representations used in this table are consistent with those described in Table~\ref{tab:main_result}.}
    \label{tab:lama2}
\end{table}

\paragraph{Can Language Models Be Used as Knowledge Bases?}
Previous work~\citep{petroni-etal-2019-language} proposes PLMs have the potential to be knowledge bases. However, \citet{cao-etal-2021-knowledgeable} questions its feasibility based on their findings that the decent performance stems from prompt bias overfitting. This paper further investigates whether PLMs can answer factual queries without overfitting from prompt bias.
As shown in the main results, we find positive evidence that PLMs can achieve relatively good performance after mitigating prompt bias, even with Manual prompts. Additionally, the performance is expected to be further improved with larger PLMs and better prompts. Therefore, we posit that~\textbf{PLMs have the potential to serve as knowledge bases. }To avoid the negative impacts of prompt bias, it is necessary to use some debiasing approach like ours in the knowledge probing process.

%% file: sec/7relatedwork.tex
\section{Related Work}
\paragraph{Factual Knowledge Extraction}
\citet{petroni-etal-2019-language} first introduce the LAMA benchmark to evaluate the factual knowledge contained in PLMs by prompting, and propose that PLMs have the potential to serve as knowledge bases.
~\citet{kassner-etal-2021-multilingual} extend the LAMA benchmark to different languages and investigate factual knowledge contained in multilingual PLMs.
One subsequent research line focuses on finding prompts with better retrieval capability.~\citet{jiang-etal-2020-how} use text-mining and paraphrasing to automatically generate prompts that show better performance than LAMA.~\citet{Shin-etal-2020-Auto} collect a training dataset consisting of different facts with LAMA and use a gradient-based searching algorithm~\citep{wallace-etal-2019-universal} to find better discrete prompts.~\citet{liu-etal-2021-gpt} and ~\citet{zhong-etal-2021-factual} take a further step by exploring continued prompt optimization on factual knowledge extraction.

However, some other works point out that PLMs can hardly serve as reliable knowledge bases currently. \citet{kassner-schutze-2020-negated} find PLMs can't distinguish negated and non-negated queries. \citet{zhong-etal-2021-factual} show that gradient-based prompts will learn patterns from the training dataset and overfit test datasets.~\citet{cao-etal-2021-knowledgeable} find PLMs suffer from prompt bias in factual knowledge extraction and propose that previous decent performance may be attributed to the prompt bias overfitting on test datasets.
Inspired by~\citet{cao-etal-2021-knowledgeable}, we take a further step to quantify the prompt bias across diverse prompts and PLMs and assess its impact on different benchmarks.  We reveal two negative impacts of prompt bias and propose a novel approach to mitigate the prompt bias in factual knowledge extraction.

There are more advanced language models~\cite{he2020deberta,zhong2022toward,zhong2023bag} for complex understanding tasks, e.g., GLUE~\cite{wang2018glue}, SuperGLUE~\cite{wang2019superglue}, in the future work, we would explore the bias and the effectiveness of our method in these models and tasks.

\paragraph{Bias in PLM}
Bias in PLMs is widely investigated in the NLP field, from training corpus~\citep{kurita-etal-2019-measuring, webster-etal-2020-measuring,dev-etal-2020-measuring} to downstream tasks such as PLM-based metrics~\citep{sun-etal-2022-BERTscore}, machine translation~\citep{stanovsky-etal-2019-evaluating, prates-etal-2020-assessing, wang-etal-2022-measuring}.
Many works explore how to mitigate the intrinsic bias in PLMs~\citep{qian-etal-2019-reducing,bordia-bowman-2019-identifying,webster-etal-2020-measuring,qianCounterfactualInferenceText2021,feiMitigatingLabelBiases2023}.
and the extrinsic bias in downstream tasks\citep{zhao-etal-2017-men, zhao-etal-2018-gender,sun-etal-2022-BERTscore, wang-etal-2022-measuring, behnke-etal-2022-bias}. Focusing on the intrinsic bias, 
counterfactual data augmentation (CDA)~\citep{zmigrodCounterfactualDataAugmentation2019,dinanQueensArePowerful2020,webster-etal-2020-measuring,barikeri-etal-2021-redditbias} involves re-balancing the training· corpus by swapping bias attribute words and taking further training. \citet{karimimahabadiEndtoEndBiasMitigation2020} and~\citet{utamaDebiasingNLUModels2020} adjust the model's training loss to mitigate bias by down-weighting the biased data in the corpus. \citet{webster-etal-2020-measuring} propose using dropout regularization~\citep{srivastavaDropoutSimpleWay} as a bias mitigation approach. \citet{schickSelfDiagnosisSelfDebiasingProposal2021} propose a post-hoc debiasing technique to discourage PLMs from generating biased text by leveraging their internal knowledge. There are also other bias mitigation technologies such as projection-based debiasing~\citep{ravfogelNullItOut2020,liangDebiasingSentenceRepresentations2020}, and contrastive learning debiasing~\citep{lyuFeatureLevelDebiasedNatural2023}. In contrast to our method, most of these approaches require the extra cost of data manipulations or model retraining. 

Our approach for performing debiasing builds on recent work that explores data-free debiasing using prompt-only querying~\citep{zhao-etal-2021-calibrate}. Different from~\citet{zhao-etal-2021-calibrate}, we mitigate bias by manipulating representation vectors instead of output probabilities. One advantage of our approach is the ability to generate debiased representation vectors, which can be used in sentence embeddings like PromptBERT~\citep{jiangPromptBERTImprovingBERT2022}. Our approach is also similar to counterfactual inference~\citep{qianCounterfactualInferenceText2021,wangShouldWeRely2022} but their approach only considers keywords in the prompt when distilling bias and discard relatively unimportant words, which have been shown can significantly affect prompts~\citep{schickExploitingClozeQuestions2021,schickItNotJust2021}; in contrast, our target is to mitigate the bias of the whole prompt.

Although generative language models~\cite{touvron2023llama,achiam2023gpt} have shown significant success in various language understanding and generation tasks~\cite{zhong2023chat,peng2023ChatGPT4MT,lu2023EAPrompt}, recent studies~\cite{zheng2023large,lyu2024beyond} show that LLMs tend to make biased choices that are inconsistent with their inherent (generated) knowledge. Our debiasing strategy consistently enhances performance and reliability on Llama2, demonstrating the universality of our method and its potential to serve as the standard post-processing toolkit for large language model.
In future work, it is also worth investigating the prompt bias in other LLM-prompting-based tasks, such as the lexical choice bias in translation~\cite{ding2020understanding}, decision bias in healthcare copilot~\cite{ren2024healthcare}.

%% file: sec/8conclusions.tex
\section{Conclusions}
In this paper, we propose a method to quantify the prompt bias in factual knowledge extraction and demonstrate two negative impacts of prompt bias: overfitting benchmarks and misleading language models. Based on these findings, we propose an approach to mitigate the prompt bias using the representation vector of prompt-only querying. Experiments show that our approach effectively rectified the inflated benchmark performance achieved by prompt bias overfitting, resulting in a more reliable evaluation of factual knowledge within PLMs. Furthermore, our approach significantly enhances the retrieval performance of prompts, even within contemporary large language models like Llama2. Additionally, we observe close performance between state-of-the-art prompts and vanilla manual prompts after mitigating prompt bias. This suggests that these ``better'' prompts achieve their performance mainly by better overfitting test datasets via prompt bias rather than their knowledge retrieval capability.
These findings contribute to the development of PLM-based knowledge base construction by shedding new light on the reliability of existing benchmarks and the actual amount of factual knowledge within PLMs.
Although our debiasing method is proposed for factual knowledge extraction, this approach can also be applied to other prompt-based tasks where the prompt bias occurs such as topic classification. For future work intending to find prompts with better retrieval capability, we strongly recommend they evaluate performance with the debiasing method.

%% file: sec/9appendix.tex
\section{Implementation Details}
\subsection{Prompt Optimization}
\label{apx:optimization}
We implement factual knowledge probing based on \huggingface~Transformers~\citep{wolf-etal-2019-tranformers} library and OpenPrompt~\citep{ding-etal-2022-openprompt} library. The OptiPrompt we reported contains 5 soft tokens and is initialized randomly. We use an AdamW~\citep{loshchilov-etal-2017-adamw} optimizer and a cosine scheduler with a warmup ratio of 0.1. We train the OptiPrompt prompts for 50 epochs with a learning rate of 3e-2 and a batch size of 16. All performance of OptiPrompt we reported is averaged results over 3 random seeds.

\subsection{Typed Querying}
\label{apx:typedquerying}
We focus on a typed querying paradigm which needs to construct a candidate set $\mathcal{C}$ for each relation. We address this problem with a few steps.

First, for each relation, we take the candidate set constructed by~\citet{kassner-etal-2021-multilingual} as our basis. Then we add labels appearing in LAMA and WIKI-UNI to the basis set, to form a more comprehensive candidate set. Next, we apply a single token filter on the set for BERT and RoBERTa, following the settings in previous work. Finally, optionally, using a common vocabulary filter on the set to fairly compare different PLMs.

The common vocabulary is obtained from the intersection of the vocabularies for different models. In our experiments, we use the common vocabulary constructed by~\citet{petroni-etal-2019-language} for BERT and the common vocabulary constructed by\citet{zhong-etal-2021-factual} for RoBERTa, which contains ~21k and ~17k case sensitive tokens respectively.

\subsection{Adjustments in the Debiasing Algorithm for Llama2}
\label{apx:llama2}
In our main experiment, the debiasing algorithm estimates prompt bias by leveraging a special [MASK] token, which is absent in the vocabulary of Llama. Therefore, we instead use "N/A"  to construct a prompt-only query.  Moreover,  considering the attention mechanism of casual language models, we make minor modifications to original prompts in cases where output slots do not occur at the prompt's end. or example, we adjust the P413 prompt from "[X] plays in [Y] position." to "[X] plays in the position of [Y]."

Another modification is that we adapt the debiasing algorithm to accommodate multi-token labels. Due to the efficacy of the Llama tokenizer, most labels in test datasets are tokenized to several tokens. For multi-token labels, we apply the debiasing strategy every time a new token is generated, dynamically updating the current prompt-only template with newly generated tokens.

\section{Additional Results on Filtered Benchmarks}
\label{apx:addition}

\begin{figure*}
\centering

\subfigure[Results on BERT-large]{
    \includegraphics[width=0.98\textwidth]{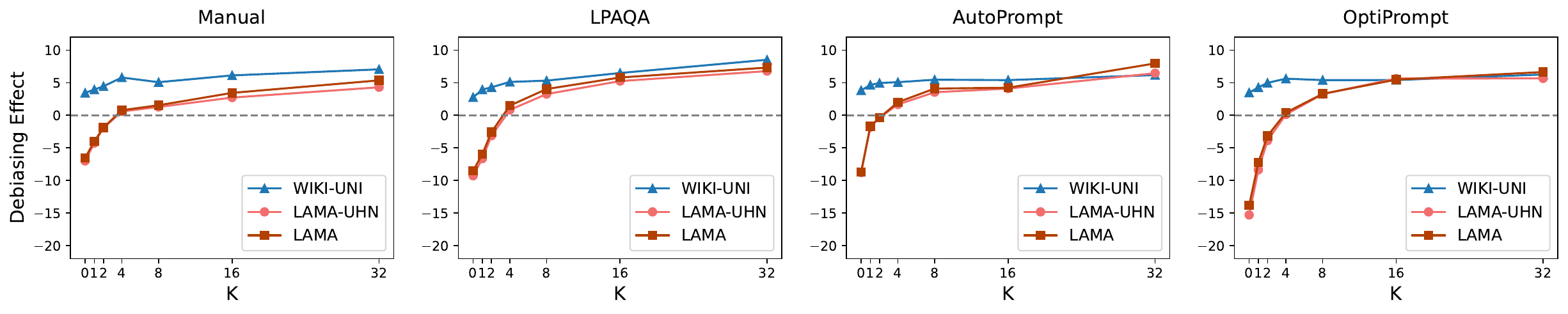}
    \label{fig:filter_out_bert-large}
}

\subfigure[Results on RoBERTa-large]{
    \includegraphics[width=0.98\textwidth]{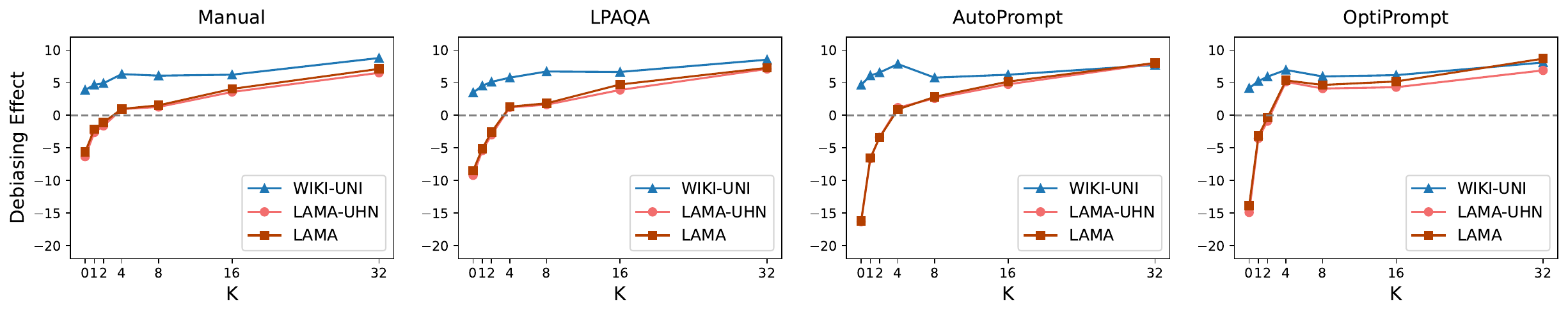}
    \label{fig:filter_out_roberta-large}
}

\caption{The impact of debiasing on benchmark accuracy after filtering out top-$k$ biased labels from datasets. We show results on the BERT-large and RoBERTa-large with various prompts and different settings of $k$.}
\label{fig:filter_out_acc_variation_othermodels}
\end{figure*}

\begin{table*}[htpb]
    \centering
    \resizebox{\linewidth}{!}{%
        \begin{tabular}{llllllll}
        \toprule
            \textbf{Prompts} & \multicolumn{7}{c}{\textbf{LAMA}} \\
            \midrule
            & \makecell[c]{\textbf{k=0}} & \makecell[c]{\textbf{k=1}} & \makecell[c]{\textbf{k=2}} & \makecell[c]{\textbf{k=4}} & \makecell[c]{\textbf{k=8}} & \makecell[c]{\textbf{k=16}} & \makecell[c]{\textbf{k=32}} \\
            \midrule
            LAMA & 100\% (34017) & 96\% (32572) & 89\% (30346) & 80\% (27366) & 73\% (24823) & 61\% (20860) & 47\% (15832) \\
            LAMA-UHN & 100\% (27102) & 96\% (25972) & 88\% (23916) & 79\% (21521) & 71\% (19326) & 60\% (16142) & 45\% (12121) \\
            WIKI-UNI & 100\% (62995) & 99\% (62451) & 95\% (59628) & 86\% (53963) & 79\% (49515) & 71\% (45015) & 63\% (39389) \\
            \midrule
            \textbf{Prompts} & \multicolumn{7}{c}{\textbf{LPAQA}} \\
            \midrule
            & \makecell[c]{\textbf{k=0}} & \makecell[c]{\textbf{k=1}} & \makecell[c]{\textbf{k=2}} & \makecell[c]{\textbf{k=4}} & \makecell[c]{\textbf{k=8}} & \makecell[c]{\textbf{k=16}} & \makecell[c]{\textbf{k=32}} \\
            \midrule
            LAMA & 100\% (34017) & 93\% (31689) & 87\% (29540) & 75\% (25470) & 68\% (23001) & 58\% (19757) & 47\% (15865) \\
            LAMA-UHN & 100\% (27102) & 93\% (25179) & 86\% (23197) & 73\% (19739) & 66\% (17963) & 56\% (15207) & 45\% (12163) \\
            WIKI-UNI & 100\% (62995) & 96\% (60171) & 95\% (59545) & 86\% (53874) & 78\% (49379) & 71\% (44766) & 62\% (39018) \\
            \midrule
            \textbf{Prompts} & \multicolumn{7}{c}{\textbf{AutoPrompt}} \\
            \midrule
            & \makecell[c]{\textbf{k=0}} & \makecell[c]{\textbf{k=1}} & \makecell[c]{\textbf{k=2}} & \makecell[c]{\textbf{k=4}} & \makecell[c]{\textbf{k=8}} & \makecell[c]{\textbf{k=16}} & \makecell[c]{\textbf{k=32}} \\
            \midrule
            LAMA & 100\% (34017) & 86\% (29193) & 79\% (26873) & 73\% (24754) & 64\% (21795) & 53\% (18011) & 42\% (14367) \\
            LAMA-UHN & 100\% (27102) & 86\% (23530) & 78\% (21307) & 69\% (18763) & 60\% (16332) & 49\% (13248) & 38\% (10256) \\
            WIKI-UNI & 100\% (62995) & 95\% (60151) & 94\% (59526) & 86\% (53871) & 79\% (49707) & 71\% (44880) & 62\% (38880) \\
            \midrule
            \textbf{Prompts} & \multicolumn{7}{c}{\textbf{OptiPrompt}} \\
            \midrule
            & \makecell[c]{\textbf{k=0}} & \makecell[c]{\textbf{k=1}} & \makecell[c]{\textbf{k=2}} & \makecell[c]{\textbf{k=4}} & \makecell[c]{\textbf{k=8}} & \makecell[c]{\textbf{k=16}} & \makecell[c]{\textbf{k=32}} \\
            \midrule
            LAMA & 100\% (34017) & 83\% (28165) & 75\% (25574) & 67\% (22638) & 56\% (19179) & 47\% (15984) & 35\% (11832) \\
            LAMA-UHN & 100\% (27102) & 81\% (21952) & 72\% (19603) & 63\% (17093) & 52\% (14213) & 43\% (11788) & 32\% (8625) \\
            WIKI-UNI & 100\% (62995) & 95\% (60148) & 91\% (57319) & 86\% (53866) & 78\% (49370) & 71\% (44697) & 62\% (38911) \\
            \bottomrule
        \end{tabular}%
    }
            \caption{Dataset sizes after filtering out top-$k$ biased labels using the prompt-only querying. We show the results of the BERT-base model with different settings of $k$. }
    \label{tab:filter_out_datasize}
\end{table*}

Figure~\ref{fig:filter_out_acc_variation_othermodels} shows the impact of debiasing after filtering out biased labels from datasets on the other two PLMs, which show a similar tendency with results in the BERT-base model shown in Figure~\ref{fig:filter_out_acc_variation}.

Table~\ref{tab:filter_out_datasize} shows the data size of three benchmarks after filtering out prompt-biased data.  The number of prompt-biased data in WIKI-UNI is much less than in LAMA and LAMA-UHN.